\documentclass{article}
\usepackage{spconf,amsmath,graphicx}

\usepackage{amssymb}
\usepackage{amsmath}
\usepackage{graphicx}
\usepackage{subfigure}
\usepackage{threeparttable}
\usepackage{multirow}
\usepackage{booktabs}

\usepackage{subfigure}
\usepackage{threeparttable}
\usepackage{multirow}
\usepackage{bm}
\usepackage{xcolor}

\definecolor{uncertain}{RGB}{255, 0, 0}

\usepackage{soul}
\definecolor{green}{RGB}{0, 100, 0}
\definecolor{red}{RGB}{139, 0, 0}
\definecolor{gold}{RGB}{255, 125, 0}
\definecolor{grey}{RGB}{200, 200, 200}
\definecolor{white}{RGB}{255, 255, 255}
\definecolor{orange}{RGB}{255, 228, 201}
\definecolor{blue}{RGB}{207, 226, 243}

\usepackage{tabularx}

\title{Comparable Demonstrations are Important in In-Context Learning: \\A Novel Perspective on Demonstration Selection}
\name{Caoyun Fan \quad Jidong Tian \quad Yitian Li \quad Hao He$^{*}$ \quad Yaohui Jin$^{*}$ \thanks{$^{*}$ Corresponding author. } \thanks{This work was supported by the Shanghai Municipal Science and Technology Major Project (2021SHZDZX0102), and the Fundamental Research Funds for the Central Universities. }}
\address{MoE Key Lab of Artificial Intelligence, AI Institute, \\Shanghai Jiao Tong University, Shanghai, China}

\begin{document}
%
\maketitle
\begin{abstract}

In-Context Learning (ICL) is an important paradigm for adapting Large Language Models (LLMs) to downstream tasks through a few demonstrations. Despite the great success of ICL, the limitation of the demonstration number may lead to demonstration bias, i.e. the input-label mapping induced by LLMs misunderstands the task's essence. Inspired by human experience, we attempt to mitigate such bias through the perspective of the inter-demonstration relationship. Specifically, we construct Comparable Demonstrations (CDs) by minimally editing the texts to flip the corresponding labels, in order to highlight the task's essence and eliminate potential spurious correlations through the inter-demonstration comparison. Through a series of experiments on CDs, we find that (1) demonstration bias does exist in LLMs, and CDs can significantly reduce such bias; (2) CDs exhibit good performance in ICL, especially in out-of-distribution scenarios. In summary, this study explores the ICL mechanisms from a novel perspective, providing a deeper insight into the demonstration selection strategy for ICL. 

\end{abstract}
\section{Introduction}
\label{s1}


Large Language Models (LLMs) \cite{DBLP:journals/tmlr/WeiTBRZBYBZMCHVLDF22} display a strong ability to perform In-Context Learning (ICL) \cite{DBLP:journals/corr/abs-2304-13712}, i.e. mastering natural language tasks from a small number of in-context demonstrations without any parameter updates \cite{DBLP:conf/acl-deelio/LiuSZDCC22,DBLP:journals/corr/abs-2305-13299}. This flexible and efficient paradigm \cite{DBLP:journals/csur/LiuYFJHN23} gives LLMs the potential to become general-purpose models \cite{DBLP:journals/corr/abs-2305-09731,Fan2023CanLL}, i.e. capable of generalizing to most tasks without further fine-tuning \cite{Fan2023ChainofThoughtTM}. 

\begin{figure}[t]
    \centering
    \subfigure[Some perspectives on exploring the ICL mechanisms. ]{
    \includegraphics[width=215pt]{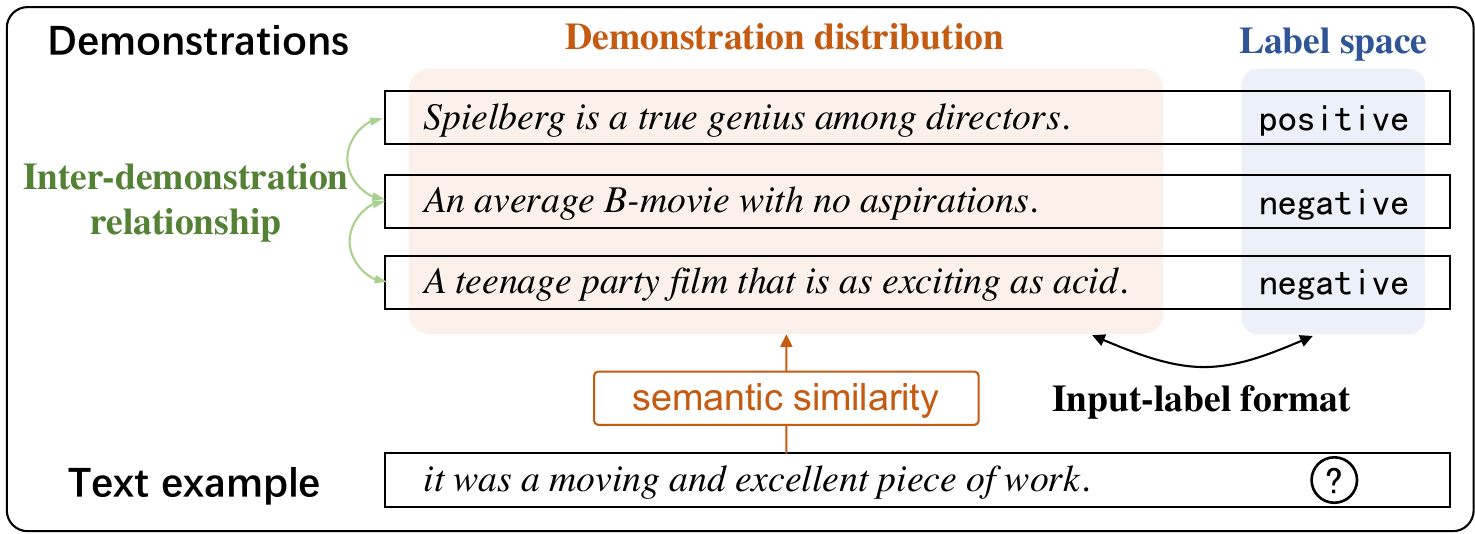} 
    \label{f1-1a}
    }
    \centering
    \subfigure[Analysis of inter-demonstration relationship. Comparable Demonstrations can reduce LLMs from misunderstanding the task's essence. ]{
    \includegraphics[width=240pt]{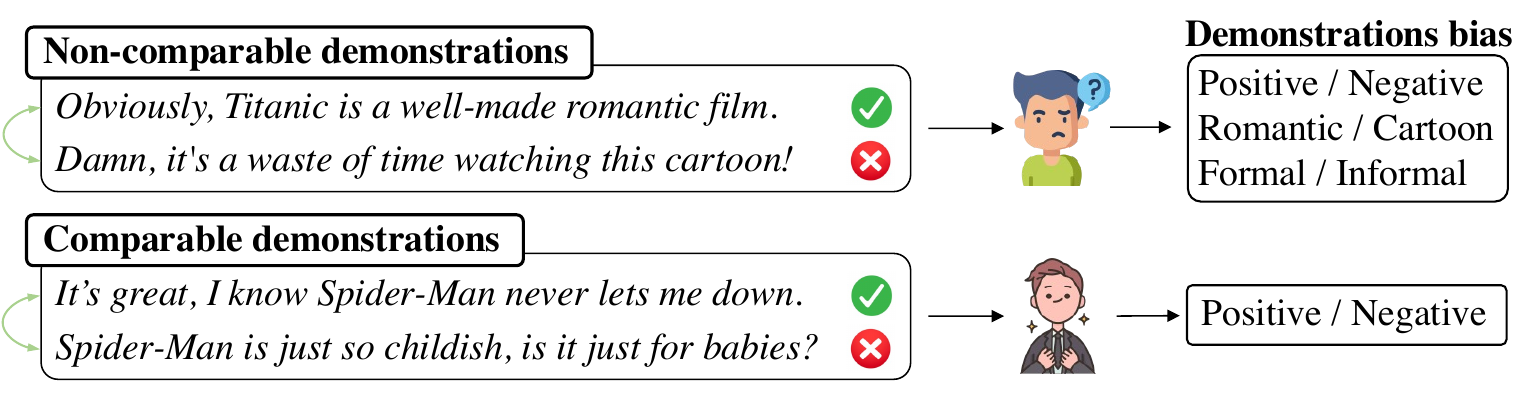} 
    \label{f1-1b}
    }
    \caption{Overview of Comparable Demonstrations in ICL. }
    \label{f1-1}
    \vspace{-0.30cm}
\end{figure}

Despite the success of ICL in many NLP scenarios, there remains little understanding of how ICL works \cite{DBLP:journals/corr/abs-2305-09731,DBLP:conf/emnlp/MinLHALHZ22}. As shown in Fig. \ref{f1-1a}, some previous studies attempted to explore the ICL mechanisms from various perspectives: \cite{DBLP:journals/corr/abs-2305-09731} considered input-label format to be important for ICL; \cite{DBLP:conf/emnlp/MinLHALHZ22,DBLP:conf/emnlp/YooKKCJLLK22} found that the label space is one of the key drivers in ICL performance; \cite{DBLP:conf/acl-deelio/LiuSZDCC22,DBLP:journals/corr/abs-2210-03493} suggested that the demonstration distribution (based on semantic similarity) can affect the information obtained by LLMs in ICL. However, as a potential perspective, the effect of the inter-demonstration relationship in ICL is not widely discussed. 

In this study, we attempt to explore the ICL mechanisms from the perspective of the inter-demonstration relationship. According to the implementation principles of ICL \cite{DBLP:conf/naacl/WebsonP22}, it requires LLMs to induce from a few demonstrations to the task's essence, i.e., a specific input-label mapping that satisfies the task. However, due to the limited number of demonstrations, the input-label mapping that conforms to the demonstrations is not unique \cite{DBLP:journals/corr/abs-2305-13299,DBLP:journals/corr/abs-2305-17256}, as shown in the top half of Fig. \ref{f1-1b}. Therefore, LLMs may induce input-label mappings that are different from the task's essence. In this study, we refer to this phenomenon as demonstration bias. 

Clearly, demonstration bias is detrimental to ICL, so how to select demonstrations to mitigate such bias is a worthwhile research direction. Based on human experience, a common solution is to consider the inter-demonstration relationship: selecting demonstrations that can be compared with each other, as shown in the bottom half of Fig. \ref{f1-1b}. In this study, we refer to such demonstrations as Comparable Demonstrations (CDs). This solution stems almost from human instinct. For example, when teaching infants to differentiate objects, humans typically select comparable demonstrations (e.g., apples and pears), rather than selecting random or identical demonstrations. Intuitively, this inter-demonstration comparison can highlight the task's essence while eliminating possible bias and spurious correlations \cite{DBLP:journals/corr/abs-2302-09345,Fan2023UnlockTP}. However, it remains unknown whether the human experience of selecting demonstrations in ICL scenarios can be applied to LLMs. 

Inspired by \cite{Kaushik2020LearningTD,Kaushik2021ExplainingTE}, we attempt to construct CDs by minimally editing the texts to flip the corresponding labels (in Section \ref{s2}). Through such manual editing, strong comparisons are created between demonstrations \cite{Yang2021ExploringTE}, thereby highlighting the task's essence. To verify the effectiveness of CDs, we conduct extensive experiments: in Section \ref{s3}, we employ instruction induction (LLMs generate descriptions of the task's essence based on demonstrations) to intuitively observe the demonstration bias of LLMs, and we confirm that CDs can significantly reduce such bias; in Section \ref{s4}, we analyze the performance of CDs in ICL scenarios from multiple perspectives, and we verify that CDs can bring performance gains to ICL, especially in the out-of-distribution scenario. 


\section{Method}
\label{s2}

\subsection{Preliminaries of In-Context Learning}

Generally, In-Context Learning (ICL) can be regarded as a conditional text generation process. LLM (parameterized by $\theta$) performs ICL with $K$ input-label pair demonstrations $D_{demo} = \left\{x_1, y_1, x_2, y_2 , \dots , x_K, y_K \right\}$, and LLM combines $D_{demo}$ to predict the label of the test example $x_t$. Specifically, this process can be represented as: 
\begin{equation}
    p(y | D_{demo}, x_t) = \prod_{t=1}^{T} p_\theta (y_t | D_{demo}, x_t, y_{<t}), 
\end{equation}

\noindent where $T$ is the generated token length and is task-specific. Here, the role of demonstrations is to help LLM elicit an input-label mapping $f: X \rightarrow Y, x \in X, y \in Y$ that is capable of matching the task's essence. 

\subsection{Comparable Demonstrations}


According to the previous analysis, due to the small number of demonstrations (e.g., $K=4$), ICL may suffer from demonstration bias. In this study, we attempt to eliminate such bias through Comparable Demonstrations (CDs). 

Specifically, the purpose of CDs is to highlight the task's essence through the inter-demonstration comparison. Inspired by Counterfactually-Augmented Data (CAD) \cite{Kaushik2020LearningTD,Kaushik2021ExplainingTE}, we can construct CDs by minimally editing the texts to flip the corresponding labels\footnote{In this study, we employ existing CAD from \cite{Kaushik2020LearningTD} to construct CDs. }. We consider that this text-level editing can maximize comparisons between demonstrations. Here, we show an example of CDs in sentiment analysis as: 
\vspace{-0.20cm}
\begin{figure}[ht]
    \centering
    \includegraphics[width=240pt]{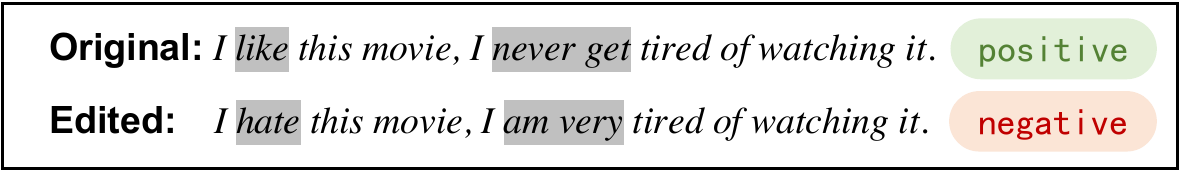}
    \label{f2-1}
\end{figure}

\vspace{-0.30cm}
\noindent This construction method has two benefits: on the one hand, the edited parts in texts are usually considered to involve the essential properties of tasks \cite{Wang2021RobustnessTS}, which helps to highlight the task's essence; on the other hand, since the majority of texts are unedited, LLMs can spontaneously eliminate potential spurious correlations in the demonstrations \cite{DBLP:journals/corr/abs-2302-09345}. 

In the fine-tuning paradigm, fine-grained manual editing (similar to CAD), as a data augmentation method, is considered inefficient and costly by the NLP community \cite{Joshi2021AnIO}. However, the ICL paradigm only requires a small number of demonstrations (e.g., $\sim$10), while the requirements for data quality are relatively high (to prevent demonstrative bias). Therefore, we consider that 
this study finds a more meaningful application scenario for fine-grained manual editing, i.e., constructing high-quality CDs for ICL. 

\section{LLM's Demonstration Bias}
\label{s3}

Based on human experience, demonstration bias is evident in ICL, but it is not yet known whether it also exists in LLMs. To explicitly observe demonstration bias in LLMs, we perform instruction induction\footnote{\cite{DBLP:journals/corr/abs-2205-10782} demonstrated that state-of-the-art LLMs (e.g. OpenAI's InstructGPT) have the ability to implement instruction induction. }: LLMs explicitly generate descriptions of the task’s essence based on a few demonstrations. By analyzing the generated instructions, we can perceive the input-label mappings induced by LLMs, and then compare them with the task's essence. We believe this is the most straightforward method to observe LLM's demonstration bias. 

\noindent \textbf{Experimental Setup:} \quad In this section, the LLM we analyzed was openAI's \texttt{gpt-3.5-turbo}, the current state-of-the-art LLM. The dataset we employed was IMDb \cite{Maas2011LearningWV}, a sentiment analysis dataset in the movie domain. The demonstration number was set to $K=4$. We tested three strategies for selecting demonstrations: random selection (random), selection based on semantic similarity (nearest), and random selection of CDs (CDs random), and each strategy generated 100 instructions. The text embeddings were obtained via openAI's \texttt{text-embedding-ada-002}, and we used cosine similarity to measure semantic similarity. Following previous research \cite{DBLP:journals/corr/abs-2205-10782}, the prompt we used is: 

\vspace{0.1cm}
\noindent\fbox{
  \parbox{235.4pt}{
  \fontsize{9.8pt}{10pt}\selectfont
    \emph{I gave a friend an instruction and some inputs. The friend read the instruction and wrote an output for every input. Here are the input-output pairs: \\ The input is $x_1$, the output is $y_1$. The input is $x_2$ \dots \\The instruction was }
}}

\vspace{0.1cm}
\noindent where $\{x\}$ refer to texts in sentiment analysis, and $y \in \left\{ \texttt{positive}, \texttt{negative} \right\}$. 



\subsection{Does Demonstration Bias Exist?}

Due to the challenging nature of the automated quality evaluation of instructions, we manually evaluated instructions generated by three demonstration selection strategies, and the evaluation results are shown in Fig. \ref{f3-1}. We find that there are significant differences between the instructions induced by LLMs and the task's essence, which implies that demonstration bias still exists in LLMs. 

To further analyze the demonstration bias in LLMs, we categorized bad instructions by error types. We summarized the error types of instructions into two categories and selected typical cases of both error types in Table \ref{tab3-1}: Case 1 highlights the detailed information in demonstrations (review of comedy), which is an over-interpretation of the input-label mapping; Case 2 misinterprets sentiment analysis as text assessment (overall evaluation), which is an over-simplification of the input-label mapping. Both error types are similar to characteristics that humans display in demonstration bias. 

\subsection{Can CDs Reduce Demonstration Bias?}

\begin{figure}[t]
    \centering
    \includegraphics[width=245pt]{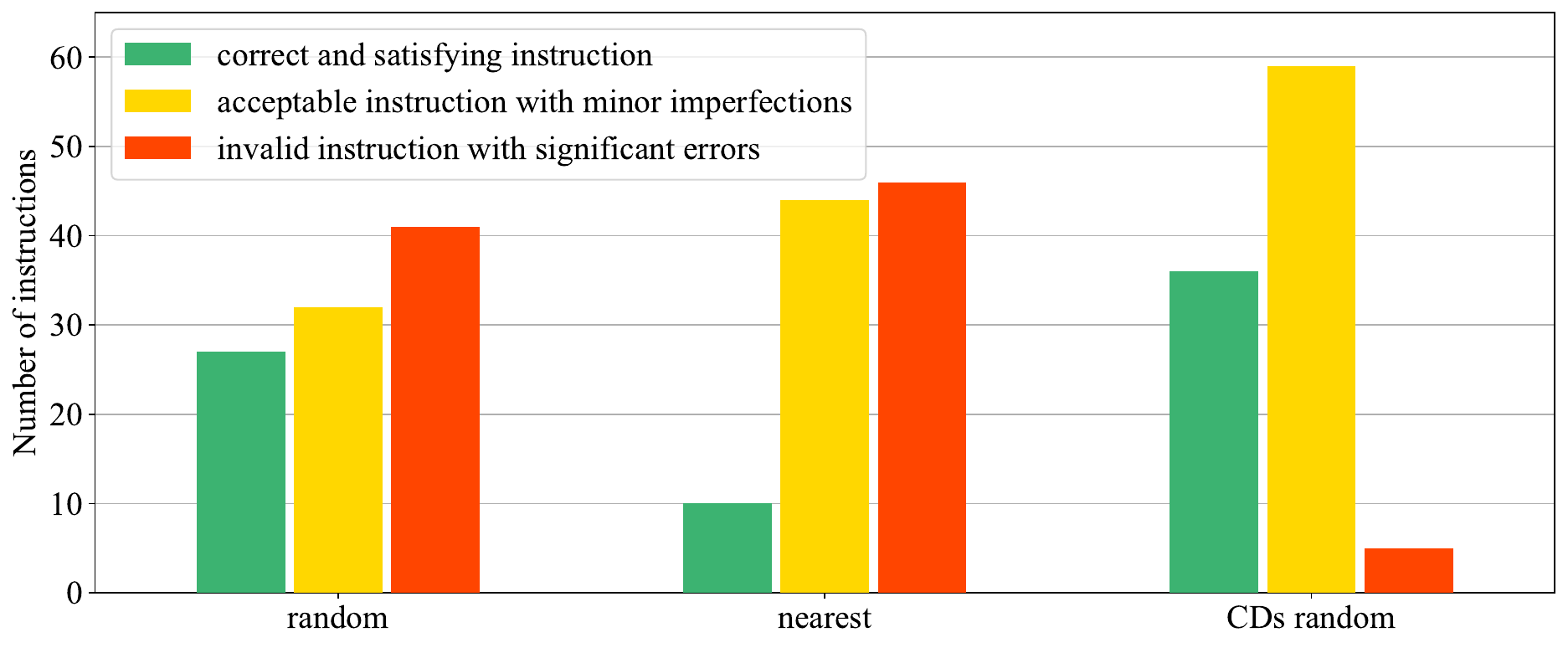}
    \caption{Comparison of instruction quality under three demonstration selection strategies. }
    \label{f3-1}
\end{figure}


As shown in Fig. \ref{f3-1}, the instructions generated by CDs significantly outperform the other two strategies, indicating the effectiveness of CDs in reducing demonstration bias. This result preliminarily indicates that the human experience of selecting demonstrations is likely to be equally effective for LLMs. In addition, it is worth noting that the performance of demonstration selection based on semantic similarity (nearest) is surprisingly poor, and we speculate that similar demonstrations contain too much repetitive information, which hinders LLMs from inducing the correct input-label mapping. 

\section{Performance of CDs in ICL}
\label{s4}

According to the previous analysis in Section \ref{s3}, it is highly likely that CDs can bring performance gains to ICL, as it has the ability to reduce demonstration bias. Therefore, we evaluate CDs in ICL scenarios and analyze the performance of CDs from multiple perspectives. 

\noindent \textbf{Experimental Setup:} \quad In this section, the LLM we analyzed was still openAI's \texttt{gpt-3.5-turbo}. Since there were existing CADs on sentiment analysis and Natural Language Inference (NLI), we conducted experiments on these two tasks. We conducted evaluations on both In-Distribution (ID) and Out-Of-Distribution (OOD) datasets: for sentiment analysis, ID dataset was IMDb \cite{Maas2011LearningWV}, and OOD datasets were Amazon review \cite{Ni2019JustifyingRU} and Yelp review \cite{Zhang2015CharacterlevelCN}; for NLI, ID dataset was SNLI \cite{Bowman2015ALA}, and OOD dataset was MNLI (split into matched and mismatched parts) \cite{Williams2018ABC}. The demonstration number was set to $K=4,8,12$. In addition to the three strategies mentioned in Section \ref{s3}, we added two more: selection based on semantic similarity in each label class (nearest class), and selection based on semantic similarity in CDs (CDs nearest). For each strategy, we sampled 500 examples in each dataset for evaluation. The text embeddings were still obtained via openAI's \texttt{text-embedding-ada-002}, and we used cosine similarity to measure semantic similarity. For sentiment analysis, the prompt we used is: \vspace{0.1cm}

\noindent\fbox{
  \parbox{235.4pt}{
  \fontsize{9.8pt}{10pt}\selectfont
    \emph{The sentence is $x_1$, the sentiment is $y_1$ $\dots$ The sentence is $x_K$, the sentiment is $y_K$. The sentence is $x_t$, the sentiment is}
}}

\vspace{0.1cm}
\noindent where $\{x\}$ refer to texts in sentiment analysis, and $y \in \left\{ \texttt{positive}, \texttt{negative} \right\}$. For NLI, the prompt we used is: 

\vspace{0.1cm}
\noindent\fbox{
  \parbox{235.4pt}{
  \fontsize{9.8pt}{10pt}\selectfont
    \emph{The premise is $x_1^p$, the hypothesis is $x_1^h$, the relation is $y_1$ $\dots$ The premise is $x_K^p$, the hypothesis is $x_K^h$, the relation is $y_K$. The premise is $x^p_t$, the hypothesis is $x^h_t$, the relation is }
}}

\vspace{0.1cm}
\noindent where $\{x^p\}$ and $\{x^h\}$ refer to premise and hypothesis in NLI, and $y \in \left\{ \texttt{entailment}, \texttt{neutral}, \texttt{contradiction} \right\}$. 

Table \ref{tab4-1} \& \ref{tab4-2} exhibit the experimental results of all demonstration selection strategies in ICL. Next, we analyze these results from three perspectives. 

\begin{table}[t]
    \centering
    \begin{threeparttable}
    \resizebox{245pt}{!}{
    \begin{tabular}{lc}
    \toprule
    \bf Type & \bf Instruction \cr
    \midrule
        Human & For each input, output whether the sentiment is positive or negative. \cr
        \midrule
        Case 1 & Read the \textcolor{red}{review of comedy} and determine if it is positive or negative. \cr
        Case 2 & Please read the following text and give an \textcolor{red}{overall evaluation}. \cr
    \bottomrule
    \end{tabular}
    }
    \end{threeparttable}
    \caption{Two types of bad instruction generated by LLMs. We mark the key parts in \textcolor{red}{red}. }  
    \label{tab3-1}
\end{table}

\begin{table*}[t]
    \centering
    \begin{threeparttable}
    \resizebox{505pt}{!}{
    \begin{tabular}{lccccccccc}
    \toprule
    \multirow{2}{*}{\bf Methods} & \multicolumn{3}{c}{\bf 4-shot} & \multicolumn{3}{c}{\bf 8-shot} & \multicolumn{3}{c}{\bf 12-shot} \cr
    \cmidrule(lr){2-4} \cmidrule(lr){5-7} \cmidrule(lr){8-10} 
     & ID & yelp & Amazon & ID & yelp & Amazon & ID & yelp & Amazon \cr
    \midrule
        random & 93.4 & 93.2 & 86.8 & 95.6 & 91.4 & 86.6 & 95.6 & 91.0 & 85.4 \cr
    \midrule
        nearest & 95.2 {\small (\textcolor{green}{+1.8\%})} & 90.4 {\small (\textcolor{red}{-2.8\%})} & 85.2 {\small (\textcolor{red}{-1.6\%})} & 95.8 {\small (\textcolor{gold}{+0.2\%})} & 86.8 {\small (\textcolor{red}{-4.6\%})} & 84.2 {\small (\textcolor{red}{-2.4\%})} & 96.3 {\small (\textcolor{gold}{+0.7\%})} & 83.8 {\small (\textcolor{red}{-7.2\%})} & 81.6 {\small (\textcolor{red}{-3.8\%})} \cr
        nearest class & \bf 95.4 {\small (\textcolor{green}{+2.0\%})} & 92.0 {\small (\textcolor{red}{-1.2\%})} & 88.4 {\small (\textcolor{green}{+1.6\%})} & \bf 96.2 {\small (\textcolor{gold}{+0.6\%})} & 91.2 {\small (\textcolor{red}{-0.2\%})} & 86.8 {\small (\textcolor{green}{-0.2\%})} & 96.3 {\small (\textcolor{gold}{+0.7\%})} & 91.0 {\small (\textcolor{gold}{+0.0\%})} & 84.4 {\small (\textcolor{red}{-1.0\%})} \cr
        \midrule
        CDs random & 93.4 {\small (\textcolor{gold}{+0.0\%})} & \bf 94.0 {\small (\textcolor{gold}{+0.8\%})} & \bf 90.0 {\small (\textcolor{green}{+3.2\%})} & 95.4 {\small (\textcolor{red}{-0.2\%})} & 93.4 {\small (\textcolor{green}{+2.0\%})} & \bf 91.4 {\small (\textcolor{green}{+4.8\%})} & 95.4 {\small (\textcolor{red}{-0.2\%})} & \bf 94.4 {\small (\textcolor{green}{+3.4\%})} & \bf 90.8 {\small (\textcolor{green}{+5.4\%})} \cr
        CDs nearest & 94.0 {\small (\textcolor{gold}{+0.6\%})} & 93.6 {\small (\textcolor{gold}{+0.4\%})} & 88.8 {\small (\textcolor{green}{+2.0\%})} & 95.8 {\small (\textcolor{gold}{+0.2\%})} & \bf 94.4 {\small (\textcolor{green}{+3.0\%})} & 89.4 {\small (\textcolor{green}{+2.8\%})} & \bf 96.4 {\small (\textcolor{gold}{+0.8\%})} & 93.8 {\small (\textcolor{green}{+2.8\%})} & 90.2 {\small (\textcolor{green}{+4.8\%})} \cr
    \bottomrule
    \end{tabular}
    }
    \end{threeparttable}
    \caption{Accuracy of different demonstration selection strategies on sentiment analysis. We consider the random selection (random) as benchmark, those with performance degradation are marked as \textcolor{red}{red}; those with performance improvement within 1\% are marked as \textcolor{gold}{yellow}; those with performance improvement above 1\% are marked as \textcolor{green}{green}. The best performance is \textbf{bold}. }
    \label{tab4-1}
\end{table*}

\begin{table*}[t]
    \centering
    \begin{threeparttable}
    \resizebox{505pt}{!}{
    \begin{tabular}{lccccccccc}
    \toprule
    \multirow{2}{*}{\bf Methods} & \multicolumn{3}{c}{\bf 4-shot} & \multicolumn{3}{c}{\bf 8-shot} & \multicolumn{3}{c}{\bf 12-shot} \cr
    \cmidrule(lr){2-4} \cmidrule(lr){5-7} \cmidrule(lr){8-10} 
     & ID & MNLI-m & MNLI-mm & ID & MNLI-m & MNLI-mm & ID & MNLI-m & MNLI-mm \cr
    \midrule
        random & 74.5 & 69.5 & 72.8 & 76.5 & 70.9 & 73.9 & 74.4 & 70.3 & 73.7 \cr
    \midrule
        nearest & \bf 75.8 {\small (\textcolor{green}{+1.3\%})} & 69.9 {\small (\textcolor{gold}{+0.4\%})} & 75.1 {\small (\textcolor{green}{+2.3\%})} & 76.8 {\small (\textcolor{gold}{+0.3\%})} & 70.2 {\small (\textcolor{red}{-0.7\%})} & 74.1 {\small (\textcolor{gold}{+0.2\%})} & 75.8 {\small (\textcolor{green}{+1.4\%})} & 71.9 {\small (\textcolor{green}{+1.6\%})} & 74.9 {\small (\textcolor{green}{+1.2\%})} \cr
        nearest class & 74.9 {\small (\textcolor{gold}{+0.4\%})} & 71.9 {\small (\textcolor{green}{+2.4\%})} & 74.5 {\small (\textcolor{green}{+1.7\%})} & 75.7 {\small (\textcolor{red}{-0.8\%})} & 73.6 {\small (\textcolor{green}{+2.7\%})} & 74.5 {\small (\textcolor{gold}{+0.6\%})} & 75.8 {\small (\textcolor{green}{+1.4\%})} & 72.6 {\small (\textcolor{green}{+2.3\%})} & 73.5 {\small (\textcolor{red}{-0.2\%})} \cr
        \midrule
        CDs random & 73.3 {\small (\textcolor{red}{-1.2\%})} & 73.0 {\small (\textcolor{green}{+3.5\%})} & 74.7 {\small (\textcolor{green}{+1.9\%})} & 75.6 {\small (\textcolor{red}{-0.9\%})} & 73.3 {\small (\textcolor{green}{+2.4\%})} & 74.0 {\small (\textcolor{gold}{+0.1\%})} & 77.2 {\small (\textcolor{green}{+2.8\%})} & 72.4 {\small (\textcolor{green}{+2.1\%})} & \bf 75.5 {\small (\textcolor{green}{+1.8\%})} \cr
        CDs nearest & 75.2 {\small (\textcolor{gold}{+0.7\%})} & \bf 73.5 {\small (\textcolor{green}{+4.0\%})} & \bf 75.3 {\small (\textcolor{green}{+2.5\%})} & \bf 77.1 {\small (\textcolor{gold}{+0.6\%})} & \bf 74.2 {\small (\textcolor{green}{+3.3\%})} & \bf 76.1 {\small (\textcolor{green}{+2.2\%})} & \bf 77.3 {\small (\textcolor{green}{+2.9\%})} & \bf 72.7 {\small (\textcolor{green}{+2.4\%})} & 74.8 {\small (\textcolor{green}{+1.1\%})} \cr
    \bottomrule
    \end{tabular}
    }
    \end{threeparttable}
    \caption{Accuracy of different demonstration selection strategies on NLI. }
    \label{tab4-2}
\end{table*}

\subsection{Dataset Distribution}

Due to the broader application scope of LLMs, we analyze the performance of CDs from the perspective of dataset distribution (ID and OOD scenarios), and our findings are: (1) strategies based on semantic similarity mainly improve ID performance, while strategies based on CDs mainly improve OOD performance; (2) the strategy that combines the advantages of both (CDs nearest) is the most competitive strategy for balancing both ID and OOD scenarios.

Our analysis of the performance differences between ID and OOD scenarios is as follows: strategies based on semantic similarity would allow LLMs to learn dataset-specific information (e.g., information about movies in IMDb dataset), which cannot be generalized to OOD scenarios; while CDs attempt to help LLMs focus on the task's essence, which to some extent alleviates the focus on dataset-specific information, thereby helping to improve the performance of LLMs in OOD scenarios. Since the application scenarios of ICL are usually uncertain, we consider that the performance gains of CDs in OOD scenarios is meaningful. 


\subsection{Task Complexity}

To clarify the extent of demonstration bias and the applicability of CDs, we analyze the experimental results from task complexity. Generally, the NLP community considers NLI to be a more complex task than sentiment analysis. 

We find that different strategies show larger performance differences on the simple task (sentiment analysis), sometimes approaching 10\%, while their performance is more stable on the complex task (NLI), typically below 3\%. This may suggest that demonstration bias is more likely to influence LLMs to induce the task's essence on simple tasks. Therefore, although the experiments show that CDs are also effective for complex tasks, they are more necessary for simple tasks. 

\subsection{Demonstration Number}

Demonstration number is often considered to affect the performance of ICL and is closely related to demonstration bias. Therefore, we analyze the performance differences between CDs and other strategies from this aspect. 

The experimental results indicate that in most cases, the performance of ICL improves as the demonstration number increases, which is in line with our expectation of demonstration bias. In addition, we find that CDs are relatively robust to the demonstration number, while strategies based on semantic similarity are more sensitive. Since the ICL paradigm is extremely sensitive to the demonstration selection, the robustness exhibited by CDs is a significant advantage. 


\section{Conclusion}

In this study, we explore the mechanisms of ICL from the perspective of the inter-demonstration relationship, and we consider that demonstration bias may exist in LLMs due to the limitation of the demonstration number. Inspired by human experience, we attempt to construct Comparable Demonstrations (CDs) by minimally editing the texts to flip the corresponding labels, to mitigate such potential bias. A series of experiments indicate that CDs bring performance gains to the ICL, especially in the OOD scenario. In the future, we aim to explore ICL in more complex scenarios (e.g. mathematical reasoning, multi-hop inference), and rigorously analyze the mechanisms of ICL from the perspective of the inter-demonstration relationship. 

We consider this study has two main limitations. First, although ICL requires relatively few demonstrations, manual annotation remains expensive, and automatically generating CDs is not currently feasible. Second, CDs only consider one-to-one relationships between demonstrations, without taking into account many-to-many relationships, which clearly does not make full use of the inter-demonstration relationship. 

\clearpage
\bibliographystyle{IEEEbib}
\bibliography{reference}

\appendix

\section{Typical Cases in Instruction Induction}
\label{a3}

In Table \ref{taba3-1}, we display more typical bad cases in instruction induction. These cases can be used as a supplement to Table \ref{tab3-1}. 

\begin{table}[ht]
    \centering
    \begin{threeparttable}
    \begin{footnotesize}
    \begin{tabularx}{219pt}{lX}
    \toprule
    \bf Type & \bf Instruction \cr
    \midrule
    Case 1 & Please write an output for each of the following inputs based on your overall impression of the movie: \textcolor{red}{positive, negative, or neutral}. \cr
    Case 2 & Read the input and write an output based on \textcolor{red}{your overall impression} of the movie. \cr
    Case 3 & Please watch this movie and give me \textcolor{red}{your honest opinion} about it. \cr
    Case 4 & Watch this \textcolor{red}{foreign film} and write your \textcolor{red}{overall impression} of it. \cr
    Case 5 & Provide \textcolor{red}{an output} for each of the given inputs. \cr
    Case 6 & Provide an output based on \textcolor{red}{your opinion of the movie or experience described in the input}. \cr
    Case 7 & Watch the movie and write a \textcolor{red}{review}. \cr
    Case 8 & Read the following \textcolor{red}{statement} and determine whether the overall sentiment is \textcolor{red}{positive, negative, or neutral}. \cr
    Case 9 & Read the \textcolor{red}{following descriptions of movies} and write 'positive' or 'negative' as the output based on \textcolor{red}{whether the reviewer liked the movie or not}. \cr
    Case 10 & Provide an output \textcolor{red}{based on your experience or opinion} of the given input. \cr
    \bottomrule
    \end{tabularx}
    \end{footnotesize}
    \end{threeparttable}
    \caption{More typical bad cases in instruction induction. Key parts are marked in \textcolor{red}{red}. }  
    \label{taba3-1}
\end{table}

These bad cases further confirm that there is also demonstration bias in LLMs. Therefore, we need to study how to eliminate this bias. 

\section{Rules for Manual Evaluation}
\label{a4}

We divide the generated instructions into three categories: 
\begin{enumerate}
    \item Correct and satisfying instruction. This refers to instructions that humans can implement 0-shot reasoning, and can generalize to out-of-distribution datasets.  
    \item Acceptable instruction with minor imperfections. This refers to the presence of some redundant words. Humans can still implement 0-shot reasoning, but cannot generalize to out-of-distribution datasets. 
    \item Invalid instruction with significant errors. This refers to the inability of even humans to implement 0-shot reasoning. The instruction completely misunderstands the essence of tasks. 
\end{enumerate}

\end{document}